# Periodic Bootstrap Thompson Sampling For Periodically Non-Stationary Bandit Problems

Boning Shao
*Faculty of Science, University of Hong Kong, Hong Kong, China*
*u3612737@connect.hku.hk*

Abstract: This paper introduces Periodic Bootstrap Thompson Sampling (PBTS), an innovative extension of the classic Thompson Sampling (TS) algorithm tailored for bandit problems with periodic non-stationarity. Conventional TS accumulates all past observations, leading to biased posteriors when reward distributions cycle over time. PBTS overcomes this by synchronizing belief resets with known or inferred period intervals and embedding structured bootstrap exploration phases, effectively purging obsolete data while preserving uncertainty estimates. PBTS is tested in artificially constructed environments, which include skewed and balanced reward distributions, along with different bootstrap proportions and misaligned periodic intervals. Results indicate that PBTS generally achieves statistically significant reductions in cumulative regret against traditional TS in periodic non-stationary environments. Subsequent discussion further articulates the potential of PBTS's real-world deployment. The study mentions limitations like extreme periodic misalignment and proposes future research such as self-adjusting cycle-recognition. With memory reset and bootstrap phase, PBTS introduces a novel approach to optimizing bandit algorithms in periodic reward contexts.

## 1 INTRODUCTION

TS is one of the most common and prevalent approaches for sequential decision-making. It is prized for its Bayesian principle to strike a balance between exploration and exploitation in multi-armed bandit scenarios. Its theoretical advantage has inspired various algorithm transformations such as integrating neural networks for high-dimensional problems (Zhang et al., 2020) and adding linearity for structured distributions (Abeille & Lazaric, 2017). However, these algorithms may not work well in cyclically non-stationary settings, where reward distributions alter periodically, like daily user activity cycles or periodic workload shifts. Standard TS continuously aggregates past data, polluting posterior estimates with outdated samples, which results in sluggish responses to distributional alternation and slow convergence to the optimal actions. Although some of TS variants can handle sudden shifts with decay factors or moving averages, they may not work in optimality in periodically shifting environments, for applications like personalized recommenders and elastic cloud computing, where predictable periodic behaviour dictates rewards.

To resolve this limitation, this paper propose PBTS algorithm. It is a novel method that considers the performance when facing periodic non-stationarity of the reward distributions. PBTS overcomes the core deficiencies of classical TS in cyclical environments through two interconnected advancements. The first innovation is periodical reset. Instead of conventional TS's persistent memory accumulation, PBTS purges obsolete data by reinitializing arms at predicted cycle boundaries. This synchronization between algorithm updates of arm parameters and environmental change of reward distributions allows swift adjustment to new reward regimes while discarding outdated historical data. Such alignment successfully prevents performance degradation from obsolete information. The second innovation is the addition of bootstrap phase at the beginning of each predicted period. Posterior resets result in initial decisions with sparse information of reward distributions at the beginning. PBTS mitigates this problem with a systematic round-robin exploration phase after the periodic resets, providing strong reward estimates before exploitation and quicker convergence in exploitation. This controlled exploration phase stabilizes the transitional period while maintaining sampling efficiency.

The introduction of these two innovations allows PBTS to perform well in periodic non-stationary environments for enhanced decision-making. Extensive testing across varied cyclical conditions attests PBTS's statistical superior regret performance relative to standard TS, with exceptional gains observed during distribution shifts where PBTS adapts almost instantaneously.

# 2 LITERATURE REVIEW

TS has gained wide reputation due to its good performance and broad applications in multi-armed bandit situations. TS was shown by Gopalan et al. to perform effectively even in challenging scenarios which include interdependent reward structures across multiple actions (Gopalan et al, 2014). Russo and Van Roy later deepened the theoretical understanding of TS algorithm through an information-theoretic framework, which links regret to the entropy of optimal-action distributions (Russo and Van Roy, 2016). There are also some algorithmic innovations like Neural TS, which integrates deep learning via neural tangent features to handle high-dimensional contexts while maintaining near-optimal regret bounds (Zhang et al., 2020). Abeille and Lazaric revisited Linear TS as a randomized optimistic algorithm, revealing its intrinsic exploration mechanism through parameter optimism (Abeille and Lazaric, 2017).

Practical adaptations demonstrate TS's cross-domain applicability. He et al. embedded an improved TS within deep Q-networks to optimize cognitive radio spectrum access (He et al., 2024). While Kong et al. applied TS to coded edge computing, reducing task allocation delays by over 15% ( Kong et al, 2025). Shi used a TS-based approach to enhance Twitter hashtag recommendation accuracy through iterative user-hashtag matrix updates (Shi, 2025). For adversarial settings, Xu et al. developed robust TS variants using pseudo-posteriors against reward poisoning attacks (Xu et al., 2024), and Uguina et al. combined TS with metaheuristics to solve dynamic team orienteering problems (Uguina et al., 2024). Other extensions include Nguyen's hyper-heuristic framework for online optimization using metaheuristic selection (Nguyen's, 2024), Wang and Tiwari's clinical trial adaptation for best treatment identification (Wang and Tiwari's, 2023), and Ferreira et al.'s revenue management system handling inventory constraints (Ferreira et al.'s , 2018).

Despite these advances, research on periodic non-stationarity remains underdeveloped. While some TS variants address general non-stationarity through sliding-window (Trovo et al., 2020) and discounted approaches (Qi et al., 2025) for abrupt or smooth changes, and specialized formulations for stochastic rising bandits (Fiandri et al., 2025), these methods lack mechanisms to exploit cyclical patterns. Existing TS variants for non-stationary settings assume stationarity or handle only abrupt or smooth shifts, neglecting cyclical patterns. They lack structured data purging or synchronized re-exploitation. The PBTS algorithm addresses this by incorporating periodic belief resets and bootstrap phases, effectively adapting to recurrent environmental dynamics within a Bayesian TS framework.

# 3 METHODOLOGY

This chapter outlines the research methodology employed in this study. Section 3.1 presents the standard TS approach, which functions as the comparative benchmark. In contrast, section 3.2 details the newly developed PBTS method, specifically designed to handle cyclical variations in reward distributions. Although both techniques employ Bayesian inference for action selection, they adopt fundamentally different strategies when responding to temporal changes in probability distributions.

## 3.1 Classic Thompson Sampling

### 3.1.1 Algorithm Procedure

```
Initialize:
  For each arm i:
    μ(i) = 0
    σ²(i) = 1
    T(i) = 0

For round t from 1 to n:
  Observe reward of arm A_t = t
  μ(A_t) <= reward
  T(A_t) <= 1
  σ²(A_t) <= S^2 / T(A_t)
For each round t > n:
  1. For each arm i:
       sample(i) = random value from Normal(μ(i), σ(i))
  2. A_t = argmax_i sample(i)
  3. Observe reward of arm A_t
  4. Update selected arm A_t:
       μ(A_t) <= (μ(A_t) * T(A_t) + reward) / (T(A_t) + 1)
```

```
            T(A_t) <= T(A_t) + 1
            σ²(A_t) <= S^2 / T(A_t)
```

Parameter explanations:

μ(i): Estimated mean reward value for option i (default 0)

$\sigma^2$(i): Measure of estimation confidence for arm i (default 1)

T(i): Selection frequency counter for arm i (default 0)

S: Predefined standard deviation of environmental noise n: Number of arms.

TS functions as an adaptive Bayesian decision-making framework that optimally navigates the exploration-exploitation trade-off. The method maintains evolving probabilistic models for each option's reward characteristics, where μ(i) captures predicted returns and $\sigma^2$(i) measures reliability of these predictions. Execution begins with standard normal prior assignments ($\mu=0$, $\sigma^2=1$, T=0) across all options, followed by mandatory initial sampling of each alternative to establish preliminary estimates. During subsequent timesteps, the algorithm samples potential rewards from the current posterior distributions and selects the arm yielding the highest sampled value. After observing the reward, it updates the belief of the reward distribution of the selected arm through Bayesian principle: the observation count T(i) increments by 1, μ(i) is updated by make a weighted average with the new reward data, and $\sigma^2$(i) is updated to $S^2/T(i)$ reflecting reduced variance of the selected arm's mean reward. The environmental noise parameter S is got by the reward distribution, where all arms follow S-sub-Gaussian distribution.

### 3.1.2 Limitations in Non-stationary Environments

The core weakness of TS in periodic non-stationary settings originates from its belief of consistent reward distributions. When reward patterns vary periodically, TS's integration of past measurements gradually pollutes present estimations with irrelevant historical information. The approach demonstrates slow adjustment to the changeable reward conditions, which is a significant drawback in cyclically varying scenarios where rapid response to reward distribution changes is crucial.

## 3.2 Periodic Bootstrap Thompson Sampling

To address the limitations mentioned in subsection 3.1.2, PBTS algorithm is proposed.

### 3.2.1 Algorithm Procedure

```
Initialize:
  Set reset period P
  Set bootstrap proportion B
  Start new period:
    For each arm i:
      μ(i) = 0
      σ(i) = 1
      T(i) = 0

sub-round t'= t % P
bootstrap step B' = (P * B // n) * n
For t' from 1 to n:
  Observe reward of arm A_t = t'
  μ(A_t) <= reward
  T(A_t) <= 1
  σ(A_t) <= S^2 / T(A_t)
For t' from n + 1 to B':
  Observe reward of arm A_t = t' % n +
1
  Update selected arm A_t:
     μ(A_t) <= (μ(A_t) * T(A_t) +
reward) / (T(A_t) + 1)
     T(A_t) <= T(A_t) + 1
     σ(A_t) <= S^2 / T(A_t)

For t' from B' + 1 to P - 1:
  1. For each arm:
       sample(i) = random value from
Normal(μ(i), σ(i))
  2. A_t = argmax_i sample(i)
  3. Observe reward of arm A_t
  4. Update selected arm A_t:
       μ(A_t) <= (μ(A_t) * T(A_t) +
reward) / (T(A_t) + 1)
       T(A_t) <= T(A_t) + 1
       σ(A_t) <= S^2 / T(A_t)
if t' = P - 1:
  Start new period
```

PBTS enhances traditional TS by introducing two components, which are periodic reinitialization at predicted reward changing point and exploratory sampling. These two mechanisms enable a strong response to recurring pattern change in data distributions. The setup is identical to traditional TS, and PBTS integrates two new parameters that manage the timing of belief resets and the duration of exploratory actions. They are the predicted cycle length P that establishes alignment points with periodic distribution shifts in the environment, and the exploration coefficient B controlling the proportion of bootstrap sampling relative to the reset intervals P. The two elements form the fundamental structural advancement while keeping the original Bayesian principles of traditional TS.

At the reset points (activated when $t \bmod P \equiv 0$), all posterior distributions return to their original non-informative states and accumulated observation records are regarded as outdated and are cleared. After each reset, a warm-up period commences, lasting approximately B×P rounds, where each action is evaluated for the same number of rounds. This bootstrapping stage guarantees adequate reward data collection before standard decision-making procedures like traditional TS. When it comes to the exploitation phase, the algorithm employs the same stochastic selection as conventional TS but relies on the data collected after the latest resetting. By cyclically filtering outdated historical information through periodic resets, and implementing deliberate bootstrap sampling phase, this approach effectively addresses the periodic alternation of the reward distributions.

### 3.2.2 Advantages

The PBTS methodology establishes an advanced approach for periodically changing environments by incorporating two novel parameters. It addresses the issue of historical data interference through cyclical resets of the arm parameters to the initial states, thereby removing old information that could otherwise hinder decision-making processes after environmental change of reward distributions occur. This systematic refresh mechanism effectively prevents the running algorithm from potentially misleading prior observations. The approach also overcomes the cold-start challenges by implementing compulsory bootstrap sampling at the beginning of the predicted intervals, where each available option receives guaranteed evaluation before transitioning to standard selection procedures. Such probing phase can yield enough reward and gain good approximations of the mean reward of all arms which are necessary for subsequent Bayesian decision processes to function more effectively. The framework's maintaining the periodic relevance and ensuring robust initialization makes it applicable for scenarios with recurring pattern changes.

# 4 EXPERIMENT

This chapter presents the comprehensive experiments designed to compare the enhanced PBTS approach with standard TS under various conditions. The investigation focuses specifically on evaluating the performance of the two algorithms in multi-armed bandit scenarios that feature cyclically changing reward patterns, which are common practical situations where traditional TS implementations may have some performance degradation. This chapter consists of three sections: the dataset utilized (4.1) introducing the data sources and their characteristics, the experiment procedure (4.2) specifying the testing methodology and implementation details, followed by the experiment results (4.3) presenting the comparative outcomes and their interpretation.

## 4.1 Data Used

This paper simulated a rating-based scenario where rewards represent integer values between 0 and 10. To model realistic non-stationary environments, synthetic data are generated for two distinct reward distribution types exhibiting periodic shifts:

Skewed Reward Environment (Type I Data): This environment consists of 10 arms with intentionally imbalanced reward distributions. The mean reward parameters are structured to create significant separation between arms. Nine subdominant arms have mean rewards uniformly sampled from [3,4.5], while one dominant arm is assigned a mean reward 5 units higher than the maximum subdominant mean. To maintain validity within the 0-10 rating scale, this dominant mean is capped at 9.5. For each arm, 1000 integer rewards between 0 and 10 are generated, ensuring through empirical validation that the actual mean of these generated rewards closely approximates the target mean parameter. The complete reward distribution for all arms undergoes regeneration at each period boundary.

Balanced Reward Environment (Type II Data): This environment also features 10 arms but without enforced performance disparity. At each period reset, all mean reward parameters are independently and uniformly resampled from the full 0 to 10 range. Similar to Type I Data, 1000 integer rewards per arm between 0 and 10 are generated, with rigorous validation ensuring the empirical mean closely matches the newly sampled target mean. This approach creates a sub-Gaussian reward structure where no arm is guaranteed dominance across periods, reflecting environments with rapidly shifting optimal options. As with the skewed environment, complete reward redistributions occur periodically.

The skewed environment captures real-world systems with inherent imbalance, such as recommendation platforms where a minority of options receive disproportionately high engagement. Conversely, the balanced environment models situations with fluid preference dynamics, such as rapidly changing consumer trends. Reward pools are made to be substantial (1000 samples per arm) to ensure statistical stability, with validation confirming that empirical means deviate from target means by less than 0.01. During periodic regenerations, all historical data are discarded before new mean parameters are sampled and fresh reward batches are generated.

## 4.2 Experiment Procedure

```
Set cumulative_regret_{1,2} = 0
For each round t:
  Get reward_1(t) from TS
  Get reward_2(t) from PBTS
  For k = 1 and 2:
    Regret_k = optimal_mean(t) - reward_k
    cumulative_regret_k <= cumulative_regret_k + regret_k
  plot cumulative_regret_{1,2} history
```

The synthetic data are utilized for running the traditional TS algorithm and the revised PBTS algorithm, calculate the instant cumulative regret and plot the regret history at every time step. Under the same circumstance, the curves from two algorithms are plotted together to conduct a comparative performance.

## 4.3 Experiment Results

Figures 1 to Figure 10 present experimental results comparing the performance of TS and PBTS across various environmental conditions. These visualizations evaluate the impact of four critical experimental variables:

1. Reward Distribution Type (Type I: Skewed, Type II: Balanced): different types of data reflect different real-world scenarios.
2. Period Length (2 cycles, 10 cycles): 2 cycles mean that the reward distributions change once, showing short-term performance, while 10 cycles mean that the reward distributions change multiple times, showing long-term performance.
3. Period Prediction Accuracy (Exact match vs. Mismatch cases): This variable considers whether the predicted period matches the actual period may influence the performance of PBTS algorithm.
4. Bootstrap Proportion B (10%, 20% of period duration): This aspect is included to consider whether the proportion of bootstrapping may have implications for the cumulative regret.

In each figure, the horizontal axis represents time steps; The vertical axis quantifies cumulative regret; Blue and yellow trajectories denote conventional TS algorithm and the proposed PBTS algorithm respectively; Vertical dashed lines indicate period boundaries where both historical data reset occurs and new reward distributions emerge.

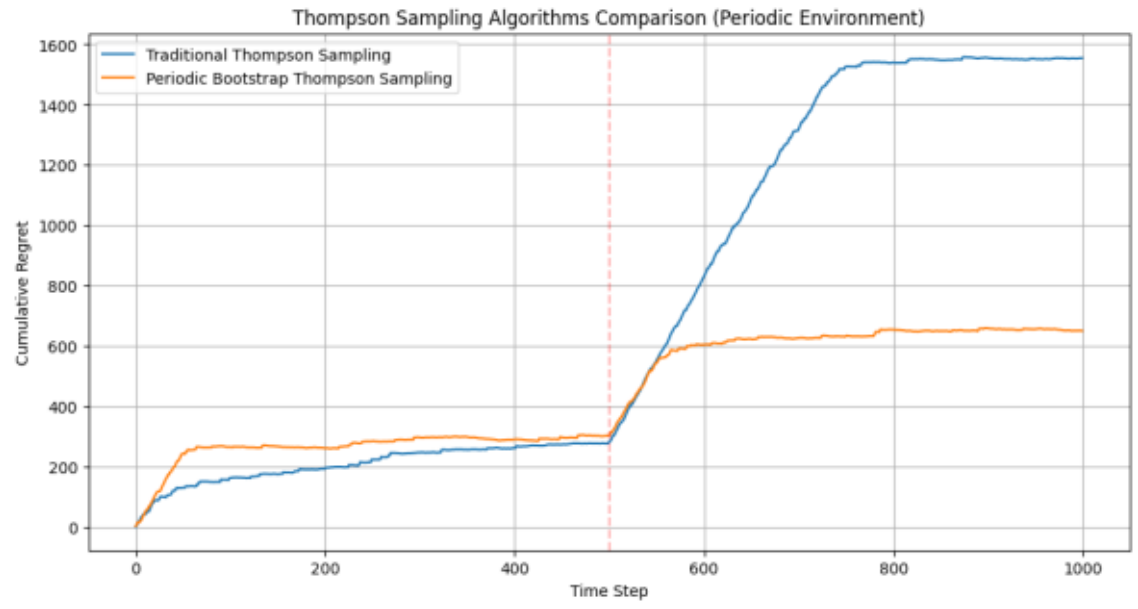

Figure 1: Type I data, 2 cycles, accurate period prediction, B = 10%. Picture credit: Original.

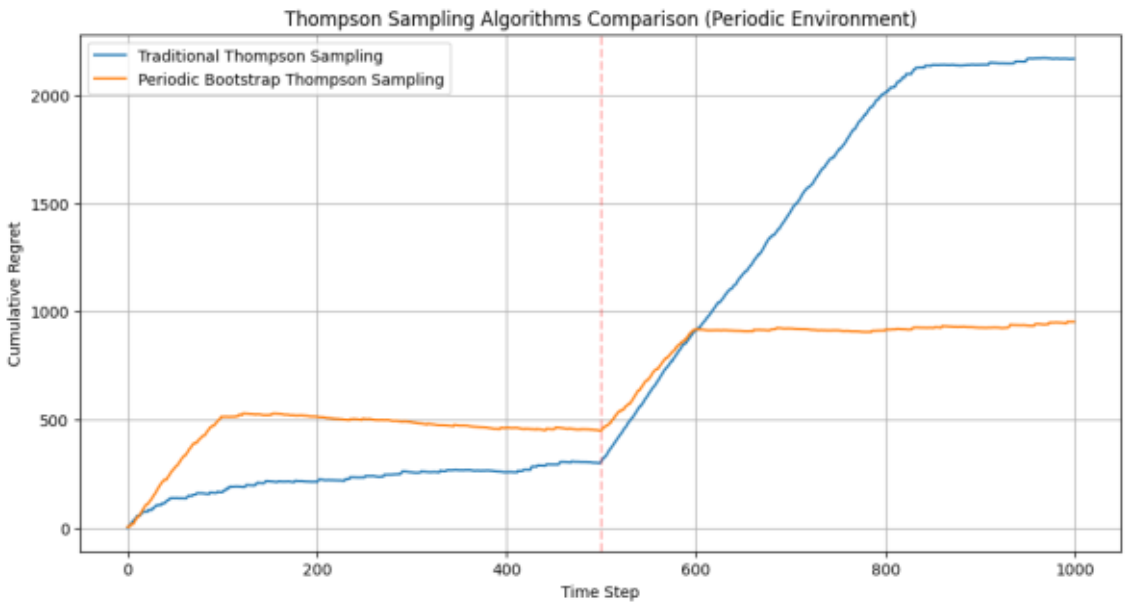

Figure 2: Type I data, 2 cycles, accurate period prediction, B = 20%. Picture credit: Original.

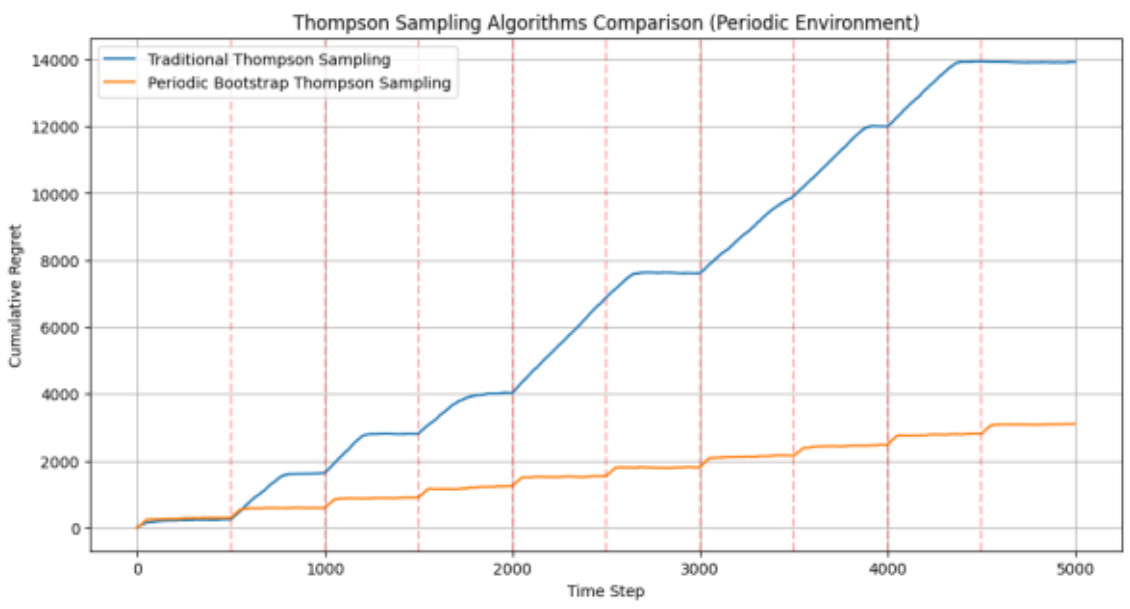

Figure 3: Type I data, 10 cycles, accurate period prediction, B = 10%. Picture credit: Original.

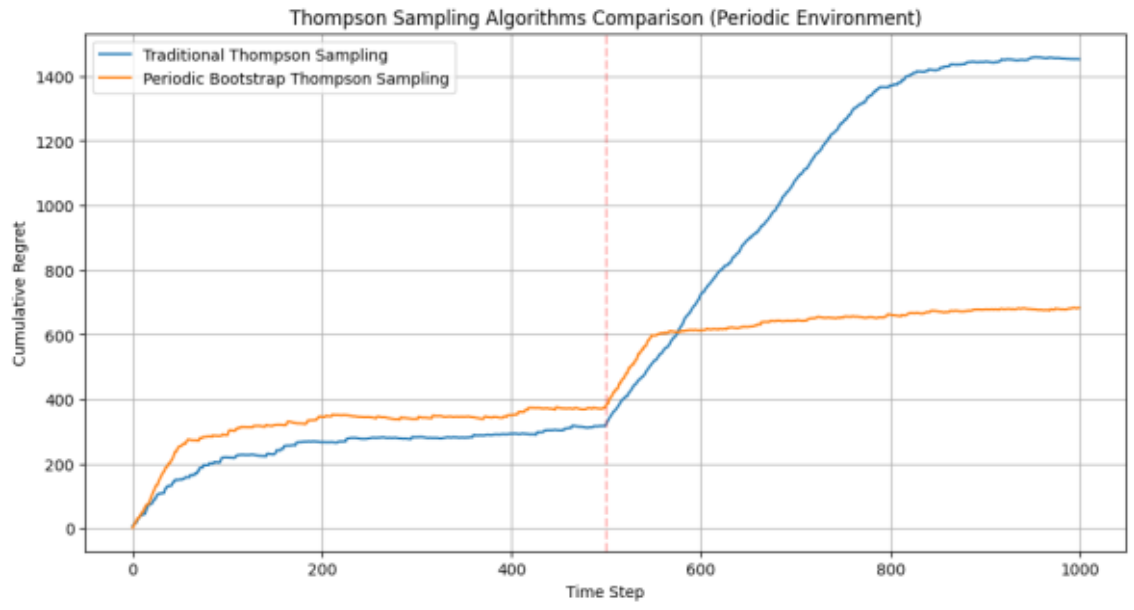

Figure 4: Type II data, 2 cycles, accurate period prediction, B = 10% (case 1). Picture credit: Original.

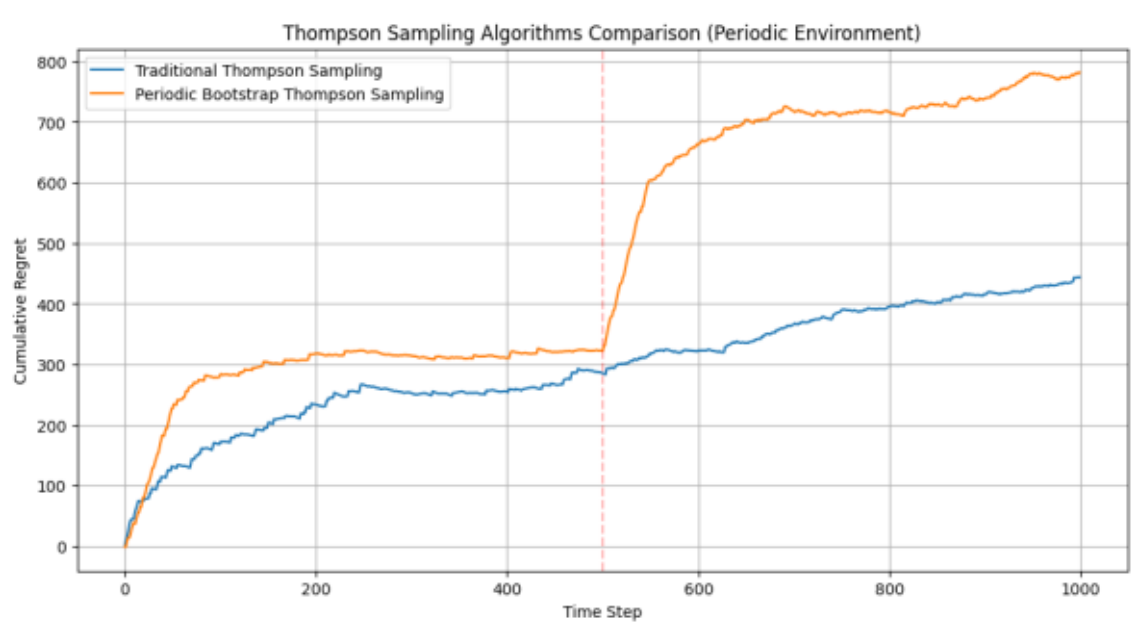

Figure 5: Type II data, 2 cycles, accurate period prediction, B = 10% (case 2). Picture credit: Original.

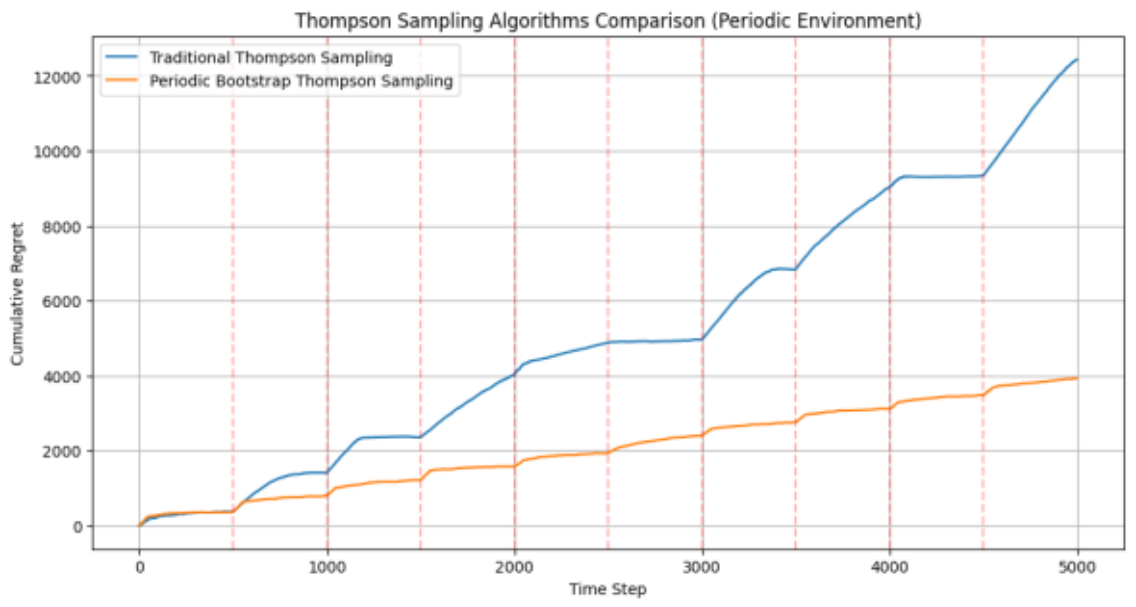

Figure 6: Type II data, 10 cycles, accurate period prediction, B = 10%. Picture credit: Original.

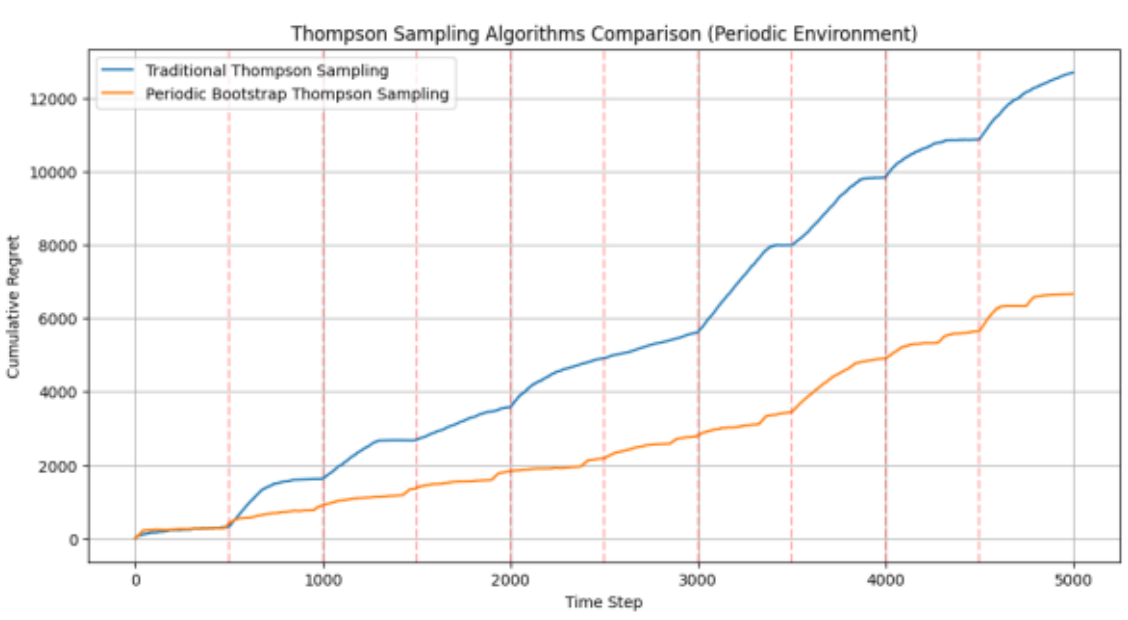

Figure 7: Type II data, 10 cycles, predicted_period = 0.95 * actual_period, B = 10%. Picture credit: Original.

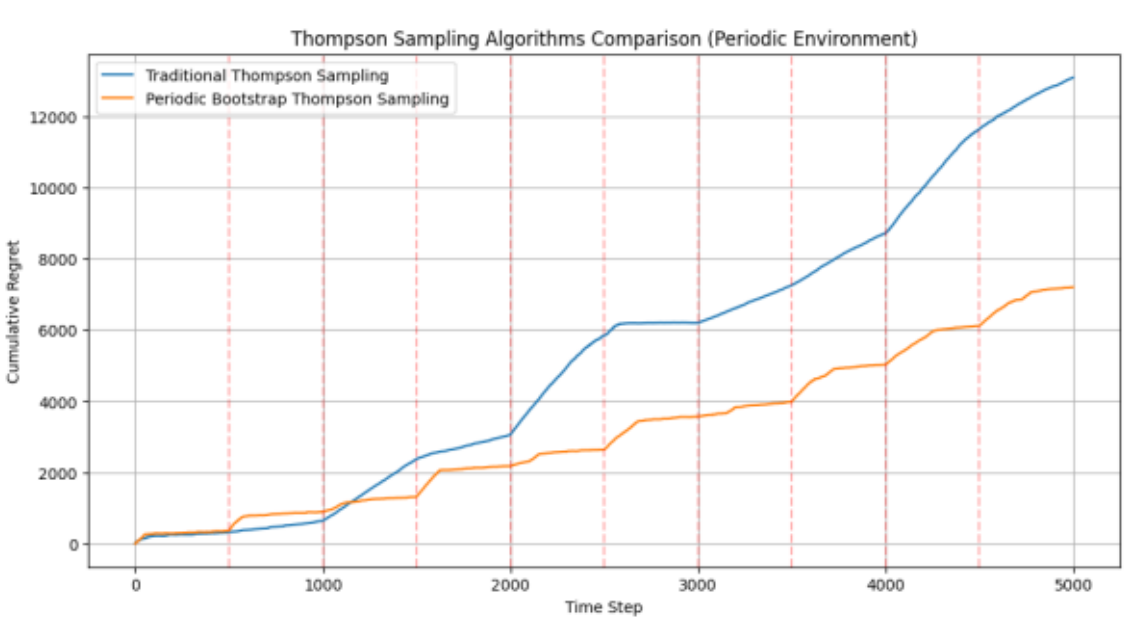

Figure 8: Type II data, 10 cycles, predicted_period = 1.05 * actual_period, B = 10%. Picture credit: Original.

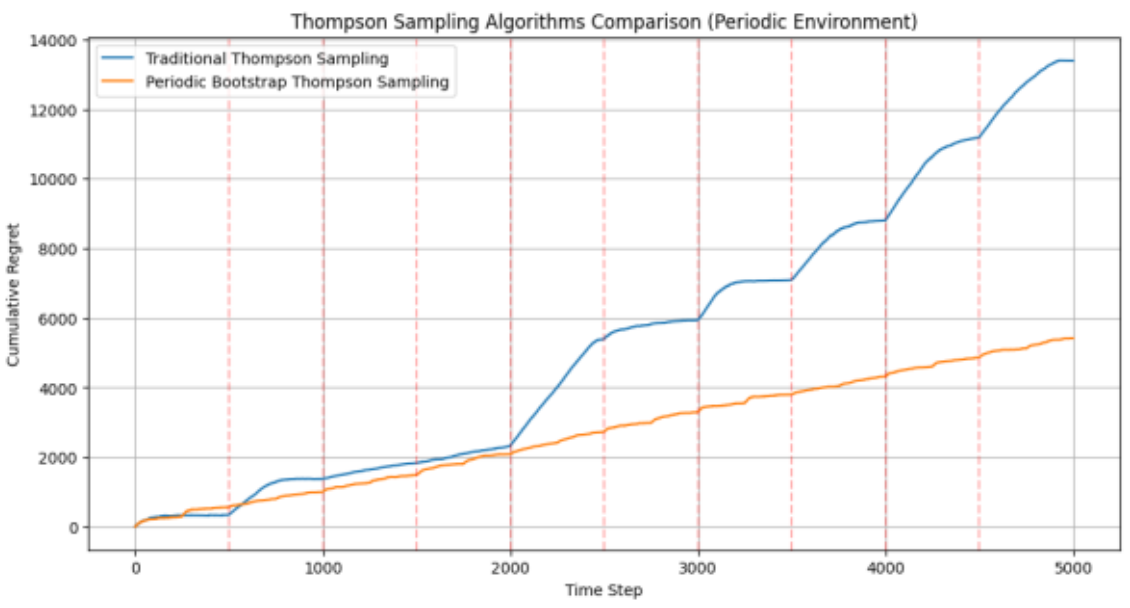

Figure 9: Type II data, 10 cycles, predicted_period = 0.5 * actual_period, B = 10%. Picture credit: Original.

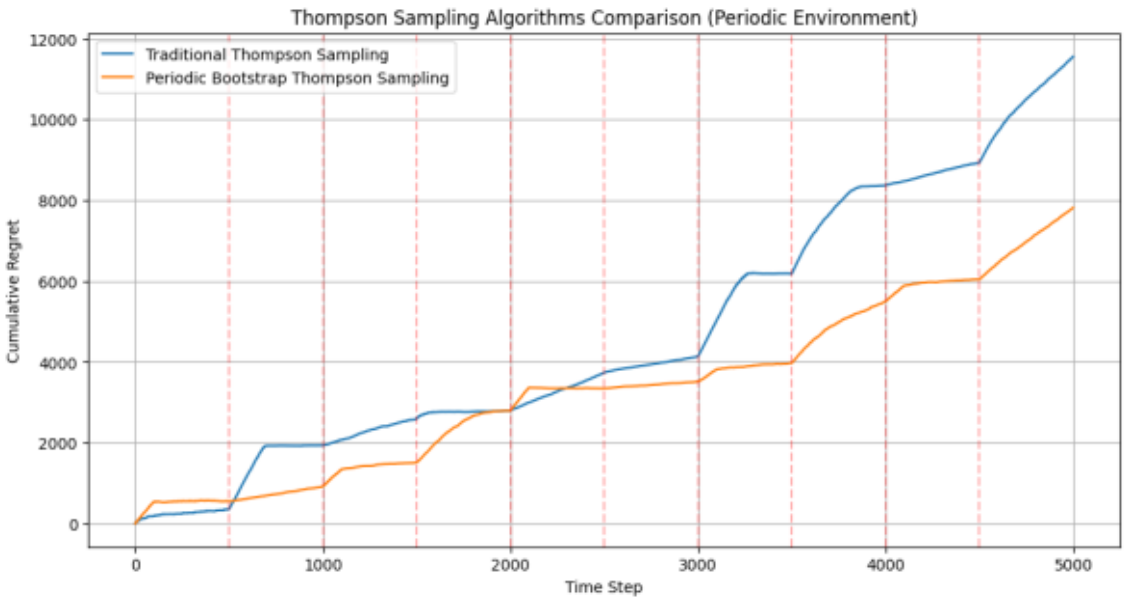

Figure 10: Type II data, 10 cycles, predicted_period = 2 * actual_period, B = 10%. Picture credit: Original.

# 5 DISCUSSION

This chapter discusses the performance and applicability of the PBTS algorithm. Experimental results show PBTS generally outperforms traditional TS in periodic non-stationary environments by reducing cumulative regret through synchronized belief resets and structured exploration. Practical applications, such as adaptive recommendation systems and cloud resource allocation, demonstrate its effectiveness in handling cyclical dynamics. This chapter entails two subsections: Experimental Analysis (section 5.1) and Applications (section 5.2).

## 5.1 Experimental Analysis

The comprehensive experimental results establish the consistent superiority of PBTS algorithm over conventional TS algorithm in periodic non-stationary environments. This performance advantage manifests across nearly all tested conditions, with PBTS demonstrating significantly lower cumulative regret throughout the evaluation timeline.

Regarding reward distribution characteristics, PBTS exhibits particularly strong performance in skewed environments (Type I data). In Figure 1, the algorithm rapidly identifies dominant arms following each reset due to its structured exploration phase, whereas conventional TS frequently becomes trapped in suboptimal exploitation cycles when transient reward spikes occur in subdominant arms. In environments with balanced reward distributions (Type II data), PBTS consistently demonstrates significant improvements in regret minimization by efficiently converging to optimal choices following distributional changes. In contrast, conventional TS approaches often struggle with insufficient and misleading exploration of the new optimal options due to the accumulating biases from historical data, as illustrated in Figure 4. There is one exception when optimal choices remain unchanged across period transitions (Figure 5). Because the optimal arm happens to be the same before and after the periodic shifts of reward distributions of all arms, PBTS may temporarily underperform compared to standard TS implementations. This situation may occur because PBTS's probing phase forces the algorithm to go through excessive exploration, which is not needed on this occasion, and PBTS also discards the previously accurate belief states. During such intervals, traditional TS benefits from accumulated knowledge to maintain effective exploitation and quick convergence, whereas PBTS accumulates some excessive regret.

Over short-term experiments covering two environmental cycles and one environmental change, PBTS effectively synchronizes its reinitialization process with distribution shifts, resulting in substantially lower regret accumulation compared to conventional TS (Figures 1, 2, and 4). While TS may show temporary advantages when environmental conditions persist across period boundaries (Figure 5), PBTS is evaluated to achieve better long-term performance across long-term experiments spanning ten cycles in both Type I and II data (Figures 3 and 6). The most notable marginal performance differences occur at distribution change points, where PBTS shows immediate adaptation while TS exhibits some response delays and slow convergence to the new optimal arm. Although standard TS maintains its advantages during stable periods, PBTS shows a long-term superiority as environmental reward distributions change continuously, demonstrating PBTS's reliable adaptation to periodic non-stationary situations.

The accuracy of period prediction may also be a crucial factor influencing the performance of PBTS. When reset intervals precisely match environmental periodicity (Figure 6), the algorithm achieves a significant reduction in cumulative regret compared to conventional TS. With moderate prediction inaccuracies involving either under- or over-estimation (Figures 7 and 8), provided that other factors do not change, PBTS shows a measurable but limited performance degradation relative to perfectly synchronized conditions. It still maintains an edge over standard TS despite these timing errors, because to some extent, the inaccurate resets can also prevent the historical data contamination that fundamentally constrains TS performance. Even under some big prediction mismatches where reset timing is greatly different from actual changing period (Figures 9 and 10), PBTS ultimately outperforms conventional TS when rounds are large by systematically avoiding complete historical belief contamination, demonstrating the robustness of this novel approach in handling periodic non-stationarity.

The bootstrap proportion parameter requires careful optimization due to its environment-dependent effects. It is necessary to determine a suitable value of the bootstrap duration to strike a balance between exploration and exploiation. In skewed reward distributions (Figures 1 and 2), increasing exploration duration provides advantages by gaining knowledge of all arms and converges faster to the optimal options within each cycle, contributing to PBTS's superior performance. However, these advantages show decreasing marginal returns and even results in marginal drawbacks because excessive exploration may lead to unnecessary sampling costs. In balanced environments (Figures 3-6), a moderate bootstrap proportions is evaluated to be effective, as extended exploration provides limited benefits while consuming resources that could otherwise support exploitation.

PBTS's periodic reset framework systematically removes outdated information that adversely affects conventional methods. Its exploration phase enables adequate exploration before exploitation begins. These two combined operations enable PBTS algorithm a rapid adaptation to periodic distributional changes rather than slow adjustment by traditional TS. These benefits are particularly evident when environmental periodicity is relatively predictable and when exploration duration matches reward structure complexity, because under these conditions, PBTS can achieve a better performance.

## 5.2 Applications

PBTS exhibits a great potential for practical implementation in real-world scenarios that feature recurring non-stationary changes of reward distributions. Its unique combination of two predetermined parameters, periodic reset and bootstrap proportion, provides measurable benefits in various contexts where traditional multi-armed bandit solutions typically fail to deliver optimal performance.

For intelligent recommendation platforms, PBTS successfully manages varying user interaction behaviours resulting from daily and seasonally cycles. When there are periodic changing cycles, PBTS can be utilized for deployment. By periodically clearing outdated engagement records while implementing the novel bootstrap sampling, the algorithm enables the system to effectively mitigate preference transition biases and converge to the new optimal actions.

The PBTS framework is also potentially applicable for distributed computing resource management, where server workloads feature daily and weekly cyclical patterns. PBTS can optimize the virtual instance deployment strategies through cyclical strategy resets and moderate exploration. The system reinitialize daily or weekly, avoiding reliance on obsolete historical information, and the exploration helps find out the optimal parameter settings.

The algorithm's applicability extends to numerous additional scenarios. It is particularly effective in situations where decision validity or data information feature periodicity of change. PBTS simply requires appropriate adjustment of its two key hyper-parameters, which are reset interval and sampling duration, to suit for specific real-world application contexts.

# 6 CONCLUSION

This work presents PBTS, an innovative solution devised to tackle the situations of recurring environmental changes through periodic resets of beliefs and bootstrap sampling strategies. The algorithm integrates two novel components into the traditional TS, reinitialization at the predicted periodic changing points of reward distributions and an exploration phase to collect data of each arm. The method successfully addresses cyclical altering patterns which traditional TS struggles to deal with. Experiments with various changing variables like data type and cycle length reveal PBTS's superior performance against conventional TS approaches across various dynamic scenarios, illustrating a great reduction in cumulative regret. The algorithm maintains stable performance under imperfect period estimation, which makes it more feasible for real-world deployment since the period prediction may not be precise enough in reality.

Several constraints of PBTS algorithm needs to be tackled and requires future research. Performance degradation may become huge when period prediction inaccuracy exceeds certain limits. To address this, self-adjusting cycle detection methods can be introduced to strengthen reliability in situations with irregular or poorly defined periodicity. Another constraint involves how to determine the appropriate exploration proportions, as ideal sampling duration vary depending on environmental features. The implementation of the method of adaptive allocation approaches responsive to reward distribution might optimize efficiency in such cases. Besides, PBTS assumes strictly regular transitions, which limits its effectiveness in scenarios where reward distributions change with an irregular interval.

Several potential directions exist to improve the performance of PBTS in future investigations. Integration of a time-based reinitialization with irregular detection capabilities may increase operational performance, allowing adaptation to hybrid environments containing both predictable and irregular changes. Self-adjusting parameter optimization systems, such as deep learning, would improve the applicability of the algorithm in unfamiliar or rapidly changing contexts and ease the requirement of work force. Selective memory of valuable information between reset cycles might speed up convergence, boosting effectiveness in situations with repeating configurations. Expansion to contextual decision-making scenarios with multiple dimensions would extend the applicability to complex real-world applications like personalized recommendation engines, where the multi-dimensional supplementary information may hugely influence outcomes.

In summary, PBTS contributes to a meaningful progression in adaptive decision-making algorithms for periodic non-stationary environments. Through resolution of current constraints in forthcoming studies, this framework can be extended to more diversified non-stationary environments, including irregular and non-cyclical changes, further solidifying PBTS as a powerful tool for complex, real-world decision-making.

# REFERENCES

Abeille, M., & Lazaric, A. (2017). Linear Thompson sampling revisited. Electronic Journal of Statistics, 11(2), 5165–5197.

Ferreira, K. J., Simchi-Levi, D., & Wang, H. (2018). Online Network Revenue Management Using Thompson Sampling. Operations Research, 66(6), 1586–1602.

Fiandri, M., Metelli, A. M., & Trovò, F. (2025). Thompson Sampling-like Algorithms for Stochastic Rising Bandits. arXiv preprint arXiv:2505.12092.

Gopalan, A., Mannor, S., & Mansour, Y. (2014). Thompson Sampling for Complex Online Problems. Proceedings of the 31st International Conference on Machine Learning, 32(1), 100-108.

He, M., Jin, M., Guo, Q., & Xu, W. (2024). Listen-After-Collision Mechanism for Dynamic Spectrum Access Using Deep Q-Network With an Improved Thompson Sampling Algorithm. IEEE Internet of Things Journal, 11(4), 6596–6606.

Kong, L., Sung, C. W., & Shum, K. W. (2025). Thompson Sampling and Proportional-Greedy Algorithm for Uncertain Coded Edge Computing. IEEE Transactions on Vehicular Technology, 74(3), 4865–4876.

Nguyen, A. (2024). Dynamic metaheuristic selection via Thompson Sampling for online optimization. Applied Soft Computing, 158, Article 111566.

Qi, H., Guo, F., & Zhu, L. (2025). Thompson Sampling for Non-Stationary Bandit Problems. Entropy, 27(1), 51.

Russo, D., & Van Roy, B. (2016). An Information-Theoretic Analysis of Thompson Sampling. Journal of Machine Learning Research, 17, 68.

Shi, Y. (2025). Research on Twitter User Tag Preference Prediction Based on Thompson Sampling Algorithm. ITM Web of Conferences, 73, 01014.

Trovo, F., Paladino, S., Restelli, M., & Gatti, N. (2020). Sliding-Window Thompson Sampling for Non-Stationary Settings. The Journal of Artificial Intelligence Research, 68, 311–364.

Uguina, A. R., Gomez, J. F., Panadero, J., Martínez-Gavara, A., & Juan, A. A. (2024). A Learnheuristic Algorithm Based on Thompson Sampling for the Heterogeneous and Dynamic Team Orienteering Problem. Mathematics, 12(11), 1758.

Wang, J., & Tiwari, R. (2023). Adaptive designs for best treatment identification with top‐two Thompson sampling and acceleration. Pharmaceutical Statistics, 22(6), 1089‐1103.

Xu, Y., Wang, Z., & Singh, G. (2024). Robust Thompson Sampling Algorithms Against Reward Poisoning Attacks. arXiv preprint arXiv:2410.19705.

Zhang, W., Zhou, D., Li, L., & Gu, Q. (2020). Neural Thompson Sampling. arXiv preprint arXiv:2010.00827.